# Localization Through Particle Filter Powered Neural Network Estimated Monocular Camera Poses

Yi Shen*[a], Hao Liu[b], Xinxin Liu[c], Wenjing Zhou[a], Chang Zhou[d], Yizhou Chen[e]

[a]University of Michigan, MI, USA ; [b]Northeastern University, Shenyang, China; [c]University of Pennsylvania, PA, USA; [d]Columbia University, NY, USA; [e]Carnegie Mellon University, PA, USA

## ABSTRACT

The reduced cost and computational and calibration requirements of monocular cameras make them ideal positioning sensors for mobile robots, albeit at the expense of any meaningful depth measurement. Solutions proposed by some scholars to this localization problem involve fusing pose estimates from convolutional neural networks (CNNs) with pose estimates from geometric constraints on motion to generate accurate predictions of robot trajectories. However, the distribution of attitude estimation based on CNN is not uniform, resulting in certain translation problems in the prediction of robot trajectories. This paper proposes improving these CNN-based pose estimates by propagating a SE(3) uniform distribution driven by a particle filter. The particles utilize the same motion model used by the CNN, while updating their weights using CNN-based estimates. The results show that while the rotational component of pose estimation does not consistently improve relative to CNN-based estimation, the translational component is significantly more accurate. This factor combined with the superior smoothness of the filtered trajectories shows that the use of particle filters significantly improves the performance of CNN-based localization algorithms.

## 1. INTRODUCTION

Recent advances in deep learning, particularly through deep neural networks (DNNs) and convolutional neural networks (CNNs), have dramatically enhanced capabilities in various domains including robotics, computer vision, and industrial automation. Neural networks have revolutionized how machines interpret complex data, enabling applications ranging from autonomous driving, and predictive maintenance[3] to advanced medical imaging[4] and environmental monitoring[5] and industrial reconstruction and matching.[6,7] These technologies not only learn from large volumes of data but also adapt to new, unforeseen scenarios without explicit programming.

In robotics and computer vision, neural networks facilitate significant improvements in object recognition, scene understanding, and decision-making processes. For instance, the implementation of progressive neural scene representations and bundle adjustments has improved localization accuracy in robotic navigation systems.[2] Similarly, enhancing visual SLAM with Bayesian filters over extended periods, Deng et al.[1] explored bolstering environmental interaction and awareness in autonomous systems. Moreover, leveraging advancements such as time-series analysis for predictive modeling, hardware-based security measures, adaptive algorithms for large street networks, and the dynamics of atmospheric electrical fields significantly contribute to the development of more robust and adaptive systems.[8–15]

Monocular cameras, highly valued for their affordability and simplicity, pose specific challenges due to their limitations in estimating the full 6-degree-of-freedom SE(3) pose without depth information. This is where neural network implementations can play a transformative role. By leveraging deep learning, it is possible to enhance the depth and pose estimation capabilities of monocular systems, as demonstrated by Ding et al.[16], who investigated the use of multiple monocular cameras for vehicle pose and shape estimation. Inspired by these advancements and the effective application of

Further author information: (Send correspondence to Yi Shen)
[a] Yi Shen: E-mail: shenrsc@umich.edu
[a] Wenjing Zhou: E-mail: wenjzh@umich.edu
[b] Liu Hao: E-mail: liuhao@stumail.neu.edu.cn
[c] Xinxin Liu: E-mail: starliu@seas.upenn.edu
[d] Chang Zhou: E-mail: mmchang042929@gmail.com
[e] Yizhou Chen: E-mail: yizhoucc@gmail.com

particle filters in vehicle localization by Liu et al.,[18] this article proposes integrating a particle filter with a CNN-based localization algorithm to further enhance SE(3) pose estimations. This initiative aims to harness the robust computational models and empirical findings from these studies.

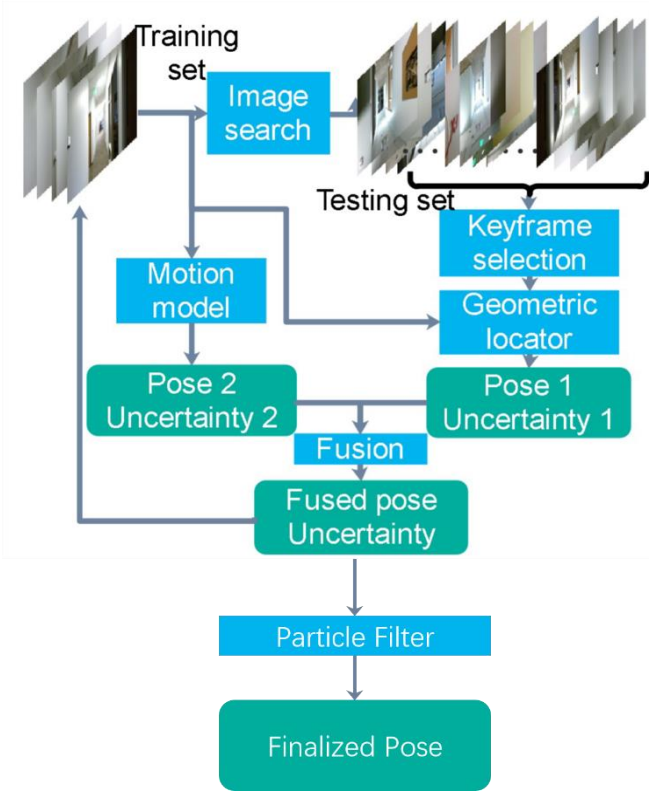

Figure 1. Flowchart of the improved system.

## 2. METHODS AND IMPLEMENTATION

The proposed system consists of two distinct components. The CNN-based Localization algorithm, which was originally produced by,[19] and the particle filter. The two will be presented separately, with the former being described just for context.

### 2.1 CNN-based Localization Algorithm

The CNN-based Localization Algorithm leverages the Siamese network framework with twin GoogleNet models trained to recognize visually similar images from a set training set.[19] This structure benefits from recent advancements in neural scene representations and local to global bundle adjustments explored by Deng et al.[2] The adaptation of smart annotation tools for point cloud data from Arief et al.[20] is considered to improve the accuracy of data labeling and processing in our algorithm. The "unfused pose estimate" is refined by an iterative process that minimizes re-projection error. We also incorporate a constant-velocity motion model to predict the camera's trajectory, enhanced with self-supervised feature tracking and depth completion to improve SLAM accuracy.[21,22] Moreover, leveraging semi-supervised learning for 3D objects,[23] bridging domain gaps in 3D scene understanding with foundation models,[24] and applying spatial-logic-aware weakly supervised learning for environmental mapping[25] further augment our methodology. Additionally, integrating advances in uncertainty quantification for deep neural networks,[26] exploiting variable correlation for anomaly detection in time series,[27] and developing Monte Carlo Tree Search algorithms for game strategy simulation[28] extend the versatility and reliability of our methodological approach. Enhancements in understanding rigid body motion dynamics also contribute to refining pose estimation methods.[29]

### 2.2 Particle Filter

Particle filters' ability to estimate states from multi-modal distributions, combined with their resiliency to noise or outliers, makes them ideal for refining the pose estimates produced by the CNN-based algorithm described above. Our particle filter is a Sample Importance Resampling (SIR) Particle Filter, relying on a Low-Variance Resampling (LVR) algorithm. The pseudocodes for these are reported in Algorithm 1 and Algorithm 2. Specifically, our SIR PF algorithm utilizes SE(3) poses for the particles, inputs, and measurements, represented as $R^{4\times 4}$ homogeneous transformation matrices, $H$, with the structure reported in (1), where $R$ is the rotation matrix, and $t$ is the position vector.

$$H = \begin{bmatrix} R & t \\ \mathbf{0}_{1\times 3} & 1 \end{bmatrix} \quad (1)$$

---
**Algorithm 1** SIR Particle Filter Algorithm
---
Require: Particles $\mathcal{X}_{k-1} = \{x_{k-1}^i, \widetilde{w}_{k-1}^i\}_{i=1}^n$, action $u_k$, measurement $z_k$, re-sampling threshold $n_t = n/3$;
1:     Initialize: particles $\{x_0^i\}_{i=1}^n$ from $p(x_0)$
2:               weights $w_0^i = 1/n$ for $i = 1, \ldots, n$.
3:     for each $x_{k-1}^i \in \mathcal{X}_{k-1}$ do
4:         $x_{k-1}^i p(x_k \mid x_{k-1}^i, u_k)$
5:         $w_k^i \leftarrow \widetilde{w}_{k-1}^i p(z_k \mid x_k^i)$
6:     end for
7:     $w_{\text{total}} \leftarrow \sum_{i=1}^n w_k^i$
8:     $\mathcal{X}_k \leftarrow \mathcal{X}_k \cup \{x_k^i, w_k^i/w_{\text{total}}\}_{i=1}^n$
9:     $n_{\text{eff}} \leftarrow 1/\sum_{i=1}^n (\widetilde{w}_k^i)^2$
10:    if $n_{\text{eff}} < n/3$ then
11:        $\mathcal{X}_k \leftarrow$ resample using $\mathcal{X}_k$
12:    end if
13:    return $\mathcal{X}_k$

---
**Algorithm 2** Low Variance Resampling Algorithm
---
Require: Particles $\mathcal{X}_k = \{x_k^i, \widetilde{w}_k^i\}_{i=1}^n$;
1:     $w_c$
       $\{\widetilde{w}_k^i\}_{i=1}^n$
2:     $r \leftarrow \text{rand}(0, n^{-1})$
3:     $j \leftarrow 1$
4:     for all $i \in \{1:n\}$ do
5:         $u \leftarrow r + (i-1)n^{-1}$
6:         while $u > w_c^j$ do
7:             $j \leftarrow j + 1$
8:         end while
9:         $x_k^i \leftarrow x_k^j$
10:       $\widetilde{w}_k^i \leftarrow n^{-1}$
11:    end for
12:    return $\mathcal{X}_k$

---

The particles in Algorithm 1 are 200 SE(3) poses, and are initialized with a uniform distribution centered on the first fused SE(3) estimate produced by the localization algorithm. The $x$, $y$, and $z$ positions of the particles are allowed to vary by ±0.25m, whilst the roll, pitch, and yaw angles ($\phi, \theta, \psi$) are allowed to vary by ±45°.

After initialization, the particles are propagated using the same constant-velocity model employed in the CNN-based algorithm from:[19] $\mathcal{C} = [\dot{x}, \dot{y}, \dot{z}, \dot{\phi}, \dot{\theta}, \dot{\psi}]^T \in \mathbb{R}^6$

Since the original control inputs were not available, $\mathcal{C}$ was instead computed by comparing sequential ground-truth poses; assuming a constant time-step of 1s, this leads to $\mathcal{C} = [\Delta x, \Delta y, \Delta z, \Delta \phi, \Delta \theta, \Delta \psi]$. To simulate uncertainty in the motion model, these perfect inputs were purposely polluted with noise, modeled as a zero-mean random distribution with covariance equal to 5% of the control input value.

To apply the input, the rotational and translational components of C are split. The rotation input, $R_{in} = R_Z R_Y R_X$, corresponding to the angular components of C is calculated via (2),[29] with rotations applied in the order $\Delta\psi, \Delta\theta, \Delta\phi$ (yaw, pitch, and then roll).

$$R_{in} = \begin{bmatrix} c_y c_p & c_y s_p s_r - s_y c_r & c_y s_p c_r + s_y s_r \\ s_y c_p & s_y s_p s_r + c_y c_r & s_y s_p c_r - c_y s_r \\ -s_p & c_p s_r & c_p c_r \end{bmatrix} \quad (2)$$

$R_{in}$ is then right-multiplied with the particle's rotation matrix, $R_{particle}$. The translation input, $t_{in} = [\Delta x, \Delta y, \Delta z]^\top$, is extracted directly from C and simply added to the particle's translation vector, $t_{particle}$. These are finally recombined into a propagated homogeneous transformation matrix, $H_{prop}$, with (3).

$$H_{prop} = \begin{bmatrix} R_{particle} R_{in} & t_{particle} + t_{in} \\ \mathbf{0}_{1\times 3} & 1 \end{bmatrix} \quad (3)$$

After propagating the particles, their weights are updated with the SE(3) pose measurements produced by the localization algorithm. This is done by multiplying their previous weights with the likelihood of recording the provided SE(3) pose measurements given the particle's own SE(3) pose. To facilitate computations, the SE(3) measurement and particle poses are both converted to their corresponding se(3) Lie Algebra (4) through the logarithmic map (5):[21]

$$se(3) = \begin{vmatrix} \hat{\omega} & v \\ 0_{1\times 3} & 0 \end{vmatrix} : \hat{\omega} = -\hat{\omega}^T \in \mathbb{R}^{3\times 3}, v \in \mathbb{R}^3 \quad (4)$$

$$\log(\cdot): SE(3) \rightarrow se(3) \quad (5)$$

The 6×1 twist coordinate vector, $[x, y, z, \phi, \theta, \psi]^T$, is then extracted from the particle's se(3) pose, and subsequently set as the mean of a multivariate normal distribution with covariance defined as a 6 × 6 identity matrix. The probability density function of this distribution, evaluated with the measurement's twist coordinate reduces the above-mentioned likelihood.

Once the particles have been re-weighted, they are normalized and the $\eta_{eff}$ parameter is calculated according to Algorithm 1. This is compared against the total number of particles divided by three to determine whether re-sampling needs to occur according to Algorithm 2.

After re-weighting, the mean and variance of the particles are computed to produce the final pose estimate for the measured pose. The particles' mean is obtained by first converting the SO(3) element of each pose to a rotation vector with (6), and extracting the translation from each pose. Following this, the mean of the first three terms of the twist coordinate vector will produce the average position of the particles. The average of each angle is instead produced by applying (7) to the last three terms of the twist coordinate vector.

$$\text{Log}(\cdot): SO(3) \rightarrow \mathbb{R}^3 \quad (6)$$

$$\alpha_{average} = \arctan 2 \left( \sum_{i=1}^{n} \cos(\alpha_i), \sum_{i=1}^{n} \sin(\alpha_i) \right), \quad \forall \alpha = \phi, \theta, \psi \quad (7)$$

To compute the variances, the zero-mean matrix $\mu_0$ is calculated by first subtracting the average twist coordinate vector from each particle's twist coordinate, and then wrapping the orientation parameters to π. Following this, the outer product of the zero-mean matrix is computed and then divided by the total number of particles to produce the unbiased covariance estimate for the measured pose (8).

$$\sigma^2 = \mu_0 \mu_0^\top / n \quad (8)$$

# 3. TESTING AND EVALUATION

## 3.1 Metrics of Assessment

To evaluate the performance of the particle filter, the unfused, fused, and filtered estimates have to be compared to the ground truth data. As all these poses are in SE(3), both the translational and rotational errors were of interest; these were computed separately and reduced to scalars using the chordal (9) and Euclidian (10) distances:

$$\Delta R = ||R_{gt}^T R_{est} - I_{3\times 3}||_F \quad (9)$$

$$\Delta t = ||t_{gt} - t_{est}||_2 \quad (10)$$

Note that $\Delta$ represents the scalar error associated with the rotational and translational components of the pose. The subscripts $_{gt}$ and $_{est}$ refer to the ground truth pose, and the estimated pose respectively. Finally, $||\cdot||_F$ refers to the Frobenius Norm. In the following section, the mean, variance, minimum, and maximum of these $\Delta t$ and $\Delta R$ errors, will be used to assess the particle filter's performance.

## 3.2 Results

Analysis with the particle filter is done on the TUM RGB-D dataset `1_floor2`.[17] Thus all of the following tables and figures will refer to this single dataset without explicitly naming it.

Table 1 below reports the minimum, maximum, mean and variance for the unfused and fused data as well as particle filter estimate errors. For the particle filter's performance, it reports the data for the upper bound and lower bound, obtained from running the filter 10 times.

Table 1. Detailed Errors Translational error in meters and Rotational error in degrees.

| Pose Error Type: | Minimum | Maximum | Mean | Variance |
|---|---|---|---|---|
| Unfused translation | 0.159 | 1.145 | 0.792 | 0.713 |
| Fused translation | 0.005 | 2.848 | 0.314 | 0.488 |
| PF upper-bound translation | 0.028 | 0.156 | 0.113 | 0.014 |
| PF lower-bound translation | 0.001 | 0.045 | 0.025 | 0.001 |
| Unfused rotation | 0.316 | 1.460 | 0.952 | 1.028 |
| Fused rotation | 0.012 | 2.293 | 0.328 | 0.397 |
| PF upper-bound rotation | 0.281 | 0.459 | 0.397 | 0.161 |
| PF lower-bound rotation | 0.055 | 0.208 | 0.137 | 0.020 |

Overall, the rotation error can either increase by up to 21%, or decrease up to 58%. The translation error is instead found to consistently decrease between 64% and 91%.

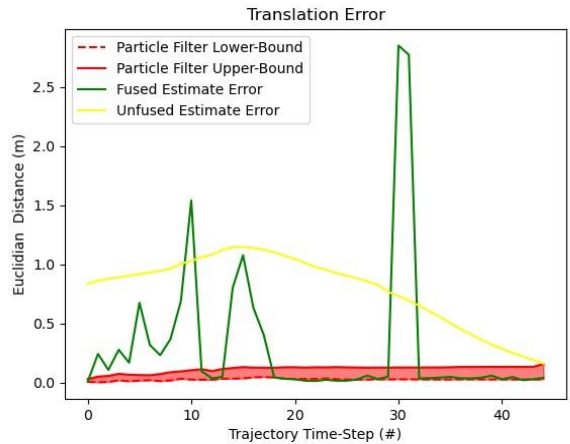

Figure 2. Translation errors of different methods.

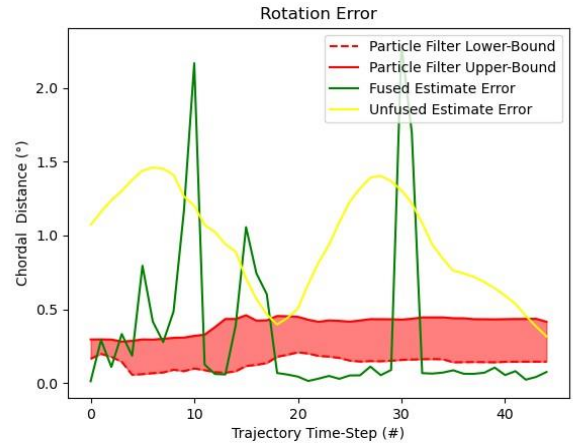

Figure 3. Rotation errors of different methods.

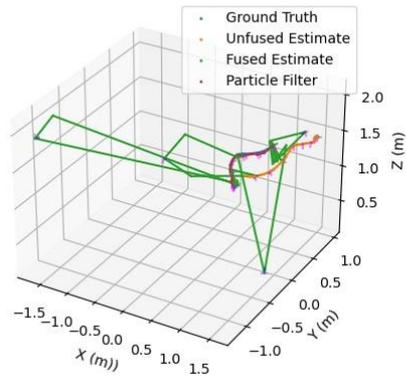

Figure 4. Trajectories of different methods, compared against the ground truth.

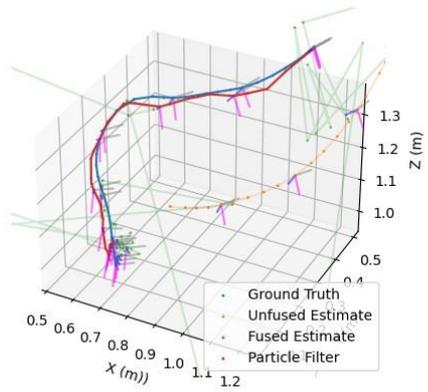

Figure 5. Zoomed trajectories of different methods, compared against the ground truth.

To visualize the data, Figure. 2 and Figure. 3 report the rotational and translational errors for the various pose estimates. Figure. 4 and Figure. 5 instead present the estimated trajectories against the ground truth. To avoid unnecessarily cluttering the plot, only a single trajectory is presented for the particle filter in this case. Additionally, the orientation is reported for every fifth pose.

## 4. DISCUSSION

### 4.1 Particle Filter Consideration

The particle filter's performance is affected by a number of parameters. First of all, both the number of particles employed and how they are initialized can have meaningful effects on the system's behavior. The choice to employ only 200 particles was made to keep computational times low and close to real-time. This did however imply that these particles had to be well initialized for the filter to converge rapidly and provide reasonable results, and is what ultimately dictated the relatively specific initialization parameters reported in Section II.B. If the number of particles was to be meaningfully increased, these could be initialized less precisely.

Another important parameter affecting the filter's performance is the covariance of the motion and measurement model. The former, whilst not explicitly provided, is believed to be a relatively accurate model because of the high degree of control the robot typically has over its motion. The latter instead had a covariance set to the relatively high value of 1 for all the 6-DoF, mostly to allow for the particles to easily converge despite the reduced precision and outliers characterizing the measurement poses.

As for the architecture of the filter, this will obviously have a major impact on its performance. The choice of an SIR particle filter was primarily dictated by their simple implementation and robustness to outliers such as the ones found in the measured poses; additionally, their adaptive re-sampling makes them computationally more efficient, enabling lower run times. As for the low-variance resampling algorithm, this was selected for its ability to rapidly allow particles to converge, as well as its potential to overall produce a more accurate estimate due to the limited variance it enforces.

Finally, the method of estimating the mean and covariance will also impact the filter's performance. The current method ignores any potential multi-modal particle distributions and assumes that these will converge sufficiently tightly so that a simple mean and covariance can be computed. Truthfully however, this might not be the case, and from this point of view the current system employs a sub-optimal solution that could be replaced by a system that considers central tendency or similar metrics.This method was also considered and tested, but it did not yield meaningfully better results and was thus abandoned.

### 4.2 Analysis of Results

The results reported in Section III.B overall suggest that the particle filter is more effective at estimating the SE(3) pose's translational components rather than its rotational counterparts. Indeed, Figure. 2 clearly shows how the Euclidian distance between the ground truth and the particle filter's estimate is significantly lower than for the estimates produced by the CNN-based localization algorithm, with an impressive average error reduction of between 64% and 91% over the entire trajectory.

Figure. 3, on the other hand, highlights how the chordal distance for the particle filter's estimate is worse in comparison to the fused estimate from the CNN-based algorithm. Indeed, Table 1 shows the minimum error for the former being meaningfully greater than that of the latter. Nonetheless, over the entire trajectory, the lack of outliers from the particle filter's estimate mitigates this effect, ultimately resulting in an average error that increases by up to 21%, or decreases by up to 58%. This variable performance could be determined by two different aspects of the particle filter: the number of particles employed how they are initialized, and how their mean and covariance are estimated. The latter aspect, as mentioned in Section IV.B, could be producing estimates that do not adequately account for multi-modal distributions of particles, or incorrectly capturing the non-linearities inherent to the orientational components of the pose. The number of particles employed, as well as their initialization, could instead be leading to sub-optimal orientation estimates if not enough particles are mapping the state-space, or if the filter becomes overly confident about its estimates and settles into local minima. This is likely what is happening since, when the particle filter was run with at least one particle initiated exactly on the first ground truth pose, it minimized the rotational error to zero.

Irrespective of this, Figure. 4 and Figure. 5 clearly show how the particle filter's trajectories are generally void of outliers and are thus considerably smoother than their CNN-derived counterparts. This is particularly important for robotic applications where severe perceptive outliers can cause the robot to take sub-optimal, unnecessary, or even unsafe actions.

## 4.3 Future Work

In light of the above results and the previously made considerations, there are three major areas of future work for the project. First, the particle filter should be evaluated over a wider range of datasets. This should be done systematically, testing its performance when the CNN-based algorithm's estimates are more or less accurate than for the currently evaluated dataset. Secondly, the algorithm should be vectorized and run with several orders of magnitude more particles. This is expected to help the particles converge better in terms of orientation, as well as potentially enabling the user to initialize the particles less precisely. Ultimately, with enough particles, we expect the system could be able to solve the kidnapped robot problem. Finally, in the future, the CNN-based localization algorithm with the particle filter could be integrated into the front-end of a graph-based SLAM algorithm, with the back-end aimed at creating small graphs that can take into account the previous poses to generate an even more refined pose estimate. Besides, since the algorithm is running on embedded systems, probably memory management[30] and hardware acceleration is an important topic as well.

## 5. CONCLUSION

Overall, the results show the particle filter generates consistently smoother trajectories than the CNN-based estimation algorithm, and outperforms the latter in terms of translational accuracy. However, the ambivalent rotational accuracy estimation, combined with the particle filter's evaluation on a single dataset, suggests that this system needs to be evaluated further before being formally recommended as a valid solution to enhance the CNN-based pose estimates.[19]